\documentclass[letterpaper, 10 pt, conference]{ieeeconf}
\IEEEoverridecommandlockouts

\usepackage{cite}
\usepackage{amsmath,amssymb,amsfonts}
\usepackage{algorithmic}
\usepackage{graphicx}
\usepackage{textcomp}
\usepackage{xcolor}
\usepackage{hyperref}
\usepackage{url}
\usepackage{todonotes}
\usepackage{booktabs} 
\usepackage{subcaption}
\usepackage{wrapfig}
\usepackage{graphicx}
\usepackage{bm}

\usepackage[font=small]{caption}

\usepackage{makecell}
\definecolor{okgreen}{HTML}{1B7F3B}
\definecolor{nored}{HTML}{C0392B}
\newcommand{\cmark}{\textcolor{okgreen}{\checkmark}}
\newcommand{\xmark}{\textcolor{nored}{\texttimes}}

\definecolor{polyumi}{HTML}{B8317A}\hypersetup{colorlinks=true,urlcolor=polyumi}

\begin{document}
\bstctlcite{IEEEexample:BSTcontrol} 

\title{\huge{\textbf{PolyUMI: Accessible Visual-Tactile-Audio Data Collection \\ for Object Inference and Manipulation}} \vspace{-0.5em}}



\author{Conor W. Hayes$^{*, 1}$, Rickmer Krohn$^{*, 2, 3, 4}$, Aravind Ramaswami$^{1}$, Anunth Ramaswami$^{1}$,\\ Nils Dengler$^{2, 3, 4}$, Kevin M. Lynch$^{1}$, J. Edward
Colgate$^{1}$, Georgia Chalvatzaki$^{2, 3, 4}$, Matthew L. Elwin$^{1}$\vspace{0.5em} \\ \href{https://polyumi-vista.github.io}{https://polyumi-vista.github.io}\vspace{-1.0em}%
\thanks{$^{*}$~Equal Contribution; $^{1}$ Center for Robotics and Biosystems, Northwestern Univ., Evanston, IL, USA}
\thanks{$^{2}$~Interactive Robot Perception \& Learning (PEARL) Lab, TU Darmstadt, Germany;
        $^{3}$ Hessian.AI;  $^{4}$ Robotics Institute Germany (RIG)}%
\thanks{Corresponding author: {\tt\small rickmer.krohn@tu-darmstadt.de}}%
}

\maketitle

\begin{abstract}
Humans typically rely on vision, touch, hearing, and proprioception to perceive contact and adapt their actions during manipulation. Providing robots with comparable responsiveness therefore requires hardware that can retain and use these complementary sensory signals. Most imitation-learning systems, however, observe demonstrations primarily through vision and proprioception, limiting access to contact information that is difficult to infer visually. We present PolyUMI, an open-source platform for scalable visual--tactile--audio demonstration collection and robot deployment. Its lightweight, wireless handheld gripper records synchronized wrist-camera, optical tactile, contact-audio, and proprioceptive observations without requiring a tethered workstation. The same sensing finger can be transferred to the robot end effector, preserving the sensing geometry between demonstration collection and policy execution. To effectively use these heterogeneous observations, we further introduce VisTA, a token-level multimodal policy that integrates information across sensors and time to predict contact-aware robot actions. Experiments spanning object inference, slip control, and contact-rich manipulation show that touch and audio reveal task-relevant information beyond vision and that VisTA is competitive with or outperforms existing multimodal policies. Together, PolyUMI and VisTA provide an accessible pipeline for collecting multimodal demonstrations and learning policies that perceive physical interaction beyond vision. Project Page: https://polyumi-vista.github.io

\end{abstract}

\section{Introduction}
\label{sec:introduction}

Learning contact-rich manipulation from demonstration requires policies that can perceive and respond to physical interaction. Here, vision already provides essential information about object pose and scene geometry, but falls short if contact events are occluded or difficult to resolve visually. Humans can instead combine vision with touch and sound to detect slip, recognize surface texture, and determine whether a component has seated correctly. Consequently, manipulation policies trained only on visual and proprioceptive observations lack direct access to these complementary cues, limiting the ability to detect and respond, e.g., to subtle changes in contact.

Prior work has begun to address this sensory gap by incorporating touch and sound into robot learning. Optical tactile sensors capture local surface deformation and contact geometry~\mbox{\cite{yuan2017gelsight,li2026simultaneous}}, while contact microphones record vibrations generated during physical interaction~\cite{liu2024maniwav}. Together with vision, these modalities have supported policies for pouring~\cite{feng_play_2024}, wrench insertion~\mbox{\cite{zhao2025polytouch}}, and dense packing~\cite{li_see_2022}. Making such capabilities broadly accessible requires practical interfaces that capture these complementary signals during human demonstrations. The Universal Manipulation Interface~(UMI)~\cite{chi2024universal} provides a foundation for this approach by enabling handheld demonstration collection without robot teleoperation. Subsequent extensions incorporate tactile sensing~\mbox{\cite{cheng2026tacumimultimodaluniversalmanipulation,zhu2025touchwildlearningfinegrained}} and contact audio~\cite{liu2024maniwav}, bringing contact information into this workflow. The remaining challenge is to integrate synchronized vision, touch, and audio into a single accessible device while maintaining consistent sensing between demonstration collection and robot deployment. This motivates a unified interface that makes multimodal demonstrations easy to collect and their sensory information directly available during policy execution.

\begin{figure}[t]
    \centering
    \includegraphics[width=\columnwidth]{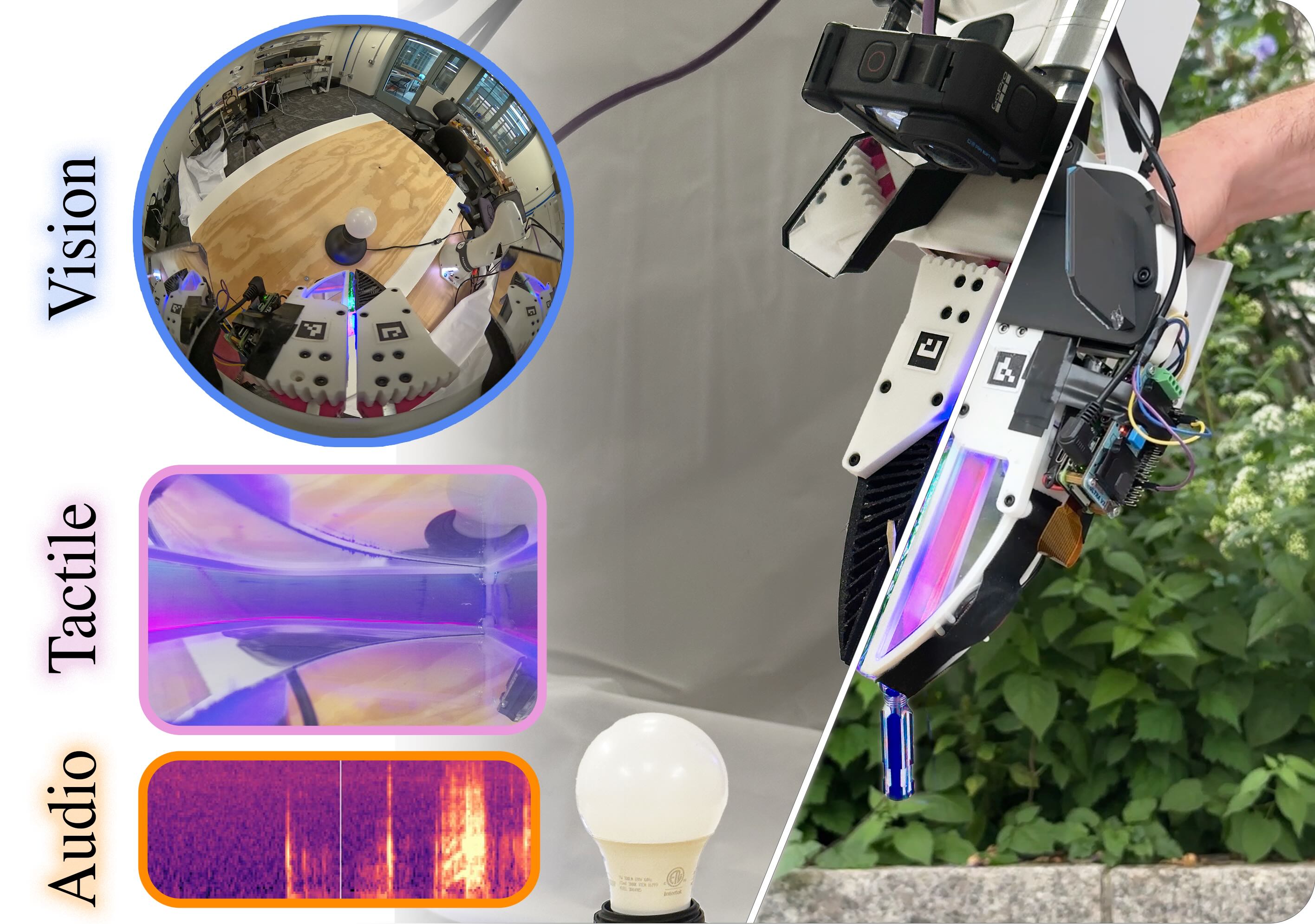}
    \caption{PolyUMI connects wireless demonstration collection (right) to robot
    policy execution (left) using a shared multimodal sensing finger. Wrist vision,
    optical tactile images, and contact audio provide complementary
    observations for learning contact-rich manipulation.}
    \label{fig:hero_figure}
    \vspace{-10px}
\end{figure}

We address this with PolyUMI, an open-source, wireless interface for synchronized visual--tactile--audio demonstration collection and robot deployment. PolyUMI combines wrist vision and proprioception with an optical tactile finger and integrated contact microphone inspired by PolyTouch~\cite{zhao2025polytouch}. The same sensing finger can be transferred between a self-contained handheld device and a robot gripper, reducing observation shifts between demonstration collection and policy execution. 
To translate these observations into robot actions, we additionally propose VisTA, a token-level multimodal policy for contact-rich manipulation. Modality-specific input stems encode the sensor observations, while a joint Transformer encoder integrates information across sensors and observation times. The fused representation conditions a flow-matching action model that predicts relative end-effector commands.

We evaluate the contribution of multimodal sensing and the effectiveness of VisTA through object-property inference, closed-loop slip control, and contact-rich manipulation. Sensor ablations show that tactile and auditory observations provide information about visually occluded object properties and support reactive slip control. In manipulation, VisTA outperforms the evaluated baselines on board wiping and matches the strongest multimodal baseline on lightbulb turning. These results highlight both the task-dependent benefits of contact sensing and the importance of combining contact feedback with visual guidance.

In summary, our contributions are:
\begin{enumerate}
    \item \textbf{PolyUMI:} an open-source, wireless platform for synchronized visual--tactile--audio demonstration collection, with a reusable sensing finger shared between handheld and robot-mounted embodiments.
    \item \textbf{VisTA:} a token-level multimodal policy that combines modality-specific observations through joint transformer fusion and a flow-matching action head.
\end{enumerate}

We release the hardware designs, electronics, firmware, fabrication instructions, and learning software on our project page: \href{https://polyumi-vista.github.io}{https://polyumi-vista.github.io}

\section{Related Work}
\label{sec:related_work}

\subsection{Multimodal Demonstration Collection}

Robot learning requires interfaces that make informative demonstrations easy to collect~\cite{chi2024universal, xu2025exumiextensiblerobotteaching, helmut_tactile-conditioned_2025, liu2025vitamin, zhaxizhuoma2024fastumi, zhu2025touchwildlearningfinegrained, li2026simultaneous}. UMI~\cite{chi2024universal} addresses this need through a portable handheld gripper and a corresponding policy-transfer framework. Touch in the Wild~\cite{zhu2025touchwildlearningfinegrained}
extends portable collection with tactile sensing and representation pretraining, while ViTaMIn~\cite{liu2025vitamin} integrates optical tactile sensing into a gripper interface. TacUMI~\cite{cheng2026tacumimultimodaluniversalmanipulation} and UMI-FT~\cite{choiIntheWildCompliantManipulation2026a} additionally incorporate force-related measurements. PolyUMI builds on these efforts by combining synchronized vision, touch, and contact audio in a wireless interface that shares the same sensing finger between collection and deployment.

\subsection{Tactile and Auditory Contact Sensing}
Touch and sound expose interaction details that can be difficult to observe with scene cameras. ManiWAV~\cite{liu2024maniwav} introduces an ear-in-hand interface for learning manipulation from in-the-wild audio-visual demonstrations. PolyTouch~\cite{zhao2025polytouch} combines optical tactile sensing, contact audio, and peripheral vision in a robotic finger for diffusion-policy learning. DIGIT~360~\cite{lambeta2024digitizingtouchartificialmultimodal} similarly integrates multiple fingertip sensing channels. Our finger follows the PolyTouch sensing principle through an independently developed, open-source implementation designed for handheld use, modular maintenance, and transfer between a robot gripper and a handheld data-collection device.

\subsection{Multimodal Policy Learning}
Exploiting complementary sensor observations requires representations that preserve their distinct information while supporting cross-modal reasoning. Visual--tactile learning has explored joint representations~\cite{lee_making_2019} and masked multimodal learning~\cite{sferrazza_power_2023}. MulSA~\cite{li_see_2022} combines vision, touch, and audio through self-attention, while stage-guided fusion~\cite{feng_play_2024} adapts modality weighting to task progress. Sparsh-X~\cite{higuera_tactile_2025} learns multisensory touch representations from tactile images, audio, motion, and pressure. VisTA builds on token-level fusion by jointly processing spatial and temporal observations to condition a flow-matching action head. Comparisons with alternative fusion methods and sensor ablations examine the contributions of sensing and policy architecture.

\begin{figure*}[t!]
  \centering
  \includegraphics[width=1.0\textwidth]{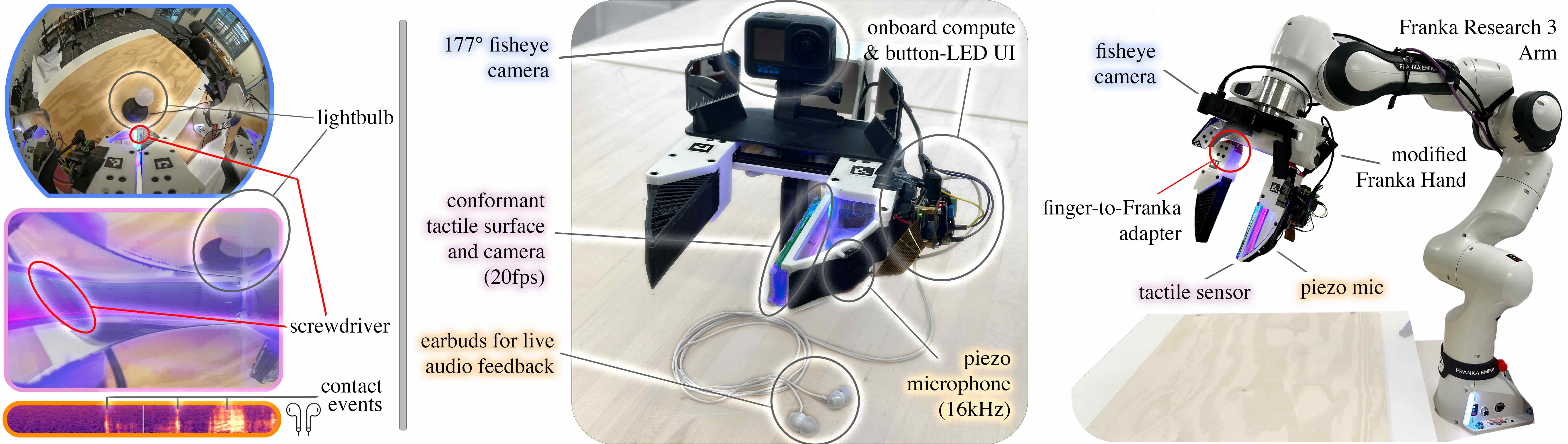}
  \caption{\textit{Left}: The relationship between different sensor views of the same scene. The tactile camera ``sees'' the screwdriver through the sensing surface, as well as the lightbulb through peripheral vision. The wrist camera sees the lightbulb directly, while the screwdriver is mostly occluded by the fingers holding it. As the device interacts with its environment, contact events are detected by the piezo microphone, which can be seen as spikes in the audio spectrogram, as well as heard through headphones during gripper operation. \textit{Center, Right}: The PolyUMI system is composed of a gripper and end-effector, carefully designed to share the same tactile finger component and task-oriented geometry.}
  \label{fig:polyumi_system}
\end{figure*}

\section{Visual-Tactile-Audio Data Collection \\ for Object Inference and Manipulation}
Contact-rich manipulation depends on information that is often occluded from vision and revealed only through deformation, vibration, and sound. 
Exploiting these signals requires reliable synchronized data collection across multiple sensors and effective multimodal fusion.
We address these challenges with PolyUMI, a wireless parallel gripper that reuses the same sensing finger across handheld demonstration collection and robot execution, providing an accessible and scalable option for multimodal data acquisition and policy deployment.
Furthermore, to effectively use PolyUMI, we propose VisTA, a token-level fusion policy that jointly reasons over visual, tactile, auditory, and proprioceptive observations to generate contact-aware actions for complex manipulation tasks.
In this section, we first describe the PolyUMI system, its modular sensing finger, and the workflow used to collect and deploy demonstrations. 
We then detail the synchronization and preprocessing of the individual sensor streams. Finally, we introduce VisTA’s observation and action representations, token-level fusion architecture, and flow-matching policy head.

\subsection{The PolyUMI System}
\label{subsec:polyumi_system}
\subsubsection{\textbf{System Overview}}
PolyUMI comprises two complementary embodiments: (1) a handheld gripper for collecting demonstrations and (2) a robot-mounted end-effector for policy execution. The handheld device follows the UMI paradigm~\cite{chi2024universal}, extended with tactile and auditory contact sensing. Both embodiments use the same transferable multimodal finger, thereby reducing observation shifts between data collection and deployment.
The handheld device includes an onboard battery, computer, and audio interface, eliminating the need for a tethered workstation and enabling practical multimodal data collection in unstructured environments. The data collection software launches automatically and is controlled through a button and status LED, while a wrist mounted camera is triggered over Bluetooth to coordinate recording across sensor streams. 
For deployment, we provide a data streaming application and a ready-to-use mounting interface for the Franka Hand in our open-source release, while the modular design allows PolyUMI to be adapted to other robotic grippers by replacing only the platform-specific mount. Table~\ref{tab:method_comparison} compares PolyUMI with existing multimodal demonstration interfaces and shows that it is the only interface that combines synchronized vision, touch, and contact audio in a handheld platform. An Overview of the proposed hardware is illustrated in Fig.~\ref{fig:polyumi_system}

\begin{table}[t]
\centering
\vspace{-0.5em}
\footnotesize
\setlength{\tabcolsep}{3pt}
\renewcommand{\cellalign}{bc}   
\begin{tabular*}{\columnwidth}{@{\extracolsep{\fill}}lcccccc@{}}
\toprule
& \multicolumn{3}{c}{Sensor modalities} & & \\
\cmidrule(lr){2-4}
Method & Vision & Tactile & Audio & \makecell{Open\\source} & \makecell{Cable\\free} & \makecell{Live\\feedback} \\
\midrule
UMI~\cite{chi2024universal}                            & \cmark & \xmark & \xmark & \cmark & \cmark & \xmark \\
FastUMI~\cite{zhaxizhuoma2024fastumi}                  & \cmark & \xmark & \xmark & \cmark & \xmark & \xmark \\
ViTaMin~\cite{liu2025vitamin}                          & \cmark & \cmark & \xmark & \xmark & \cmark & \xmark \\
Actuated UMI~\cite{helmut_tactile-conditioned_2025}    & \cmark & \cmark & \xmark & \cmark & \cmark & \xmark \\
ManiWAV~\cite{liu2024maniwav}                          & \cmark & \xmark & \cmark & \cmark & \cmark & \xmark \\
TacThru~\cite{li2026simultaneous}                      & \cmark & \cmark & \xmark & \cmark & \xmark & \xmark \\
Touch in the Wild~\cite{zhu2025touchwildlearningfinegrained} & \cmark & \cmark & \xmark & \cmark & \xmark & \cmark \\
\textbf{PolyUMI} (Ours)                                & \cmark & \cmark & \cmark & \cmark & \cmark & \cmark \\
\bottomrule
\end{tabular*}
\caption{Comparison of multimodal data collection systems. PolyUMI is the only interface able to collect synchronous vision, tactile and audio data. 
It also supplies a novel modality of operator feedback through live audio monitoring. The system's hardware and software are open-sourced for community use.}
\label{tab:method_comparison}
\vspace{-1.0em}
\end{table}

\subsubsection{\textbf{Modular Multimodal Sensing Finger}}
The multimodal finger integrates an optical tactile sensor and a contact microphone into a compact module. Its sensing principle is inspired by PolyTouch~\cite{zhao2025polytouch}. However, since the original implementation is not publicly available, we independently developed the mechanical structure, electronics, firmware, and fabrication process, and adopted several alternative design choices which can be seen in the open source release, to make the sensor softer and easier to manufacture. We also added dynamic lighting compensation to enable the use of the finger across multiple environments, necessary for an in-the-wild data collector. 
The output specifications of PolyUMI's modalities are summarized in Table~\ref{tab:sensors}. In the following we will explain each part of the system in more detail.

\paragraph{\textbf{Optical Tactile Sensing}}
An internal camera observes a deformable reflective surface through a curved mirror, providing a near-overhead view of a large contact region. In comparison to PolyTouch (which uses 3 layers), our sensing surface consists of seven layers of VHB tape coated with aluminum powder and sealed with medical tape. The layered construction increases conformability, while the reflective coating makes local deformation visible to the internal camera. The tactile stream is recorded at 20\,fps with a resolution of $1152\times648$. Camera parameters are adjusted online to maintain usable observations under changes in ambient illumination during in-the-wild data collection.

\paragraph{\textbf{Contact Audio}}
A contact microphone is rigidly coupled to the sensing-finger. In contrast to an external microphone, it primarily records vibrations transmitted through the finger and grasped object. These vibrations provide temporally precise cues for events such as contact onset, sliding, impact, and slip~\cite{liu2024maniwav}. Audio is sampled as a 16\,kHz mono PCM signal through a Raspiaudio ULTRA+ interface. A headphone output additionally allows the demonstrator to monitor the contact-audio signal while collecting data, which led to more consistent demonstrations during contact-phases. 

\begin{table}[t]
\centering
\vspace{-0.5em}
\footnotesize
\setlength{\tabcolsep}{4pt}
\renewcommand{\arraystretch}{1.15}
\begin{tabular*}{\columnwidth}{@{\extracolsep{\fill}}lll@{}}
\toprule
Modality & Sensor & Output \\
\midrule
\addlinespace[2pt]
Tactile        & Optical finger  & 20\,fps, $1152\times648$ MJPEG \\
Audio      & Contact mic & 16\,kHz mono PCM \\
Vision         & GoPro Hero 12 + fisheye     & 60\,fps, $1920\times1080$ MP4 \\
Proprioception & SLAM / ArUco           & 6-DoF pose + gripper width \\
\bottomrule
\end{tabular*}
\vspace{-5px}
\caption{PolyUMI sensing modalities and their native output specifications.}
\label{tab:sensors}
\vspace{-1.5em}
\end{table}

\paragraph{\textbf{Wrist Vision}}
PolyUMI uses a GoPro Hero~12 with a MAX Lens Mod~2.0, providing an approximately $177^{\circ}$ field of view. Side mirrors expose additional views of the gripper and manipulated object within the egocentric image. The camera records $1920\times1080$ video at 60\,fps. During preprocessing, the video is decoded and sampled at timestamps aligned with the policy observations.

\paragraph{\textbf{Proprioception}}
The handheld and robot-mounted embodiments obtain proprioception differently but express it in a common end-effector representation. For handheld demonstrations, monocular-inertial ORB-SLAM3~\cite{ORBSLAM3_TRO} estimates the six-degree-of-freedom gripper trajectory from the GoPro video and IMU (via a custom fork which supports more precise gripper masking during mapping). ArUco markers attached to the fingers provide the gripper width. During robot execution, the end-effector pose is computed from the robot joint state, while the gripper width is read directly from the hand.

\subsubsection{\textbf{Manufacturing and Assembly}}

PolyUMI is designed for straightforward manufacturing, assembly, and maintenance. As summarized in Table~\ref{tab:cost_time_comparison}, adding the multimodal sensing finger to the original UMI platform requires $\approx \$240$ in additional hardware and four hours of assembly. The complete design uses modular components and minimal software dependencies, reducing the setup effort required to deploy the system.
Importantly, the finger can be easily transferred between the gripper and the robot so that policies can be trained and deployed on a single sensor, minimizing domain shift. Transferring the finger between the two embodiments takes $\approx 10$ minutes.

\begin{table}[t]
    \centering
    \footnotesize
    \setlength{\tabcolsep}{4pt}
    \renewcommand{\arraystretch}{1.15}
    \begin{tabular*}{\columnwidth}
        {@{\extracolsep{\fill}}lccc@{}}
        \toprule
        & \makecell{Robot\\End-Effector~\cite{chi2024universal}}
        & \makecell{Handheld\\UMI Gripper~\cite{chi2024universal}}
        & \makecell{Multimodal\\Sensing Finger} \\
        \midrule
        Hardware cost & \$742.48 & \$700.27 & \$235.96 \\
        Assembly time & 30\,min & 2\,h & 4\,h \\
        \bottomrule
    \end{tabular*}
    \caption{Estimated hardware cost and manual assembly time for the main
    PolyUMI components. The multimodal sensing finger adds approximately
    \$240 and four hours of assembly relative to the original UMI platform.}
    \label{tab:cost_time_comparison}
     \vspace{-1.0em}
\end{table}

\subsubsection{\textbf{Synchronization and Data Processing}}
\label{subsec:data_processing}

PolyUMI's sensors operate at different sampling rates and latencies. We associate all observations with a common trajectory timeline and construct policy inputs at a 10\,Hz control frequency. All observations are interpolated to the timestamp of the newest observation from the stream with the highest latency. Additionally, the first few actions in each chunk are skipped in execution, to compensate for the total combined latency from the synchronized observations to to the action execution. The exact number of actions skipped is calculated through a mix of online measurement and offline calibration. At each policy step, the policy receives the current and previous wrist-camera and tactile images, resulting in an observation history of~$H=2$. Both image modalities are resized to $224\times224$ pixels before being provided to the model.
The continuous 16\,kHz audio signal is divided into windows aligned with the policy observations.
Each window is converted into a log-Mel spectrogram using 128 Mel bands, a 25\,ms analysis window, and a 10\,ms hop. We retain 48 spectrogram frames, corresponding to approximately~0.5\,s of recent contact audio. The resulting audio observation has shape $3\times128\times48$.
This processing produces aligned visual, tactile, auditory, and proprioceptive observations while retaining the temporal resolution required to capture short contact events.

\subsubsection{\textbf{Control and Execution}}
Following UMI~\cite{chi2024universal}, each forward pass of the policy outputs a sequence of $\mathrm{SE}(3)$ positions in gripper frame. On the arm, these 10\,Hz commands are interpolated and translated into joint torques at 1\,kHz by a cartesian impedance controller (adapted from \cite{luo_serl_2025}) to achieve compliant manipulation. To achieve precise gripper control, the commercially available Franka Hand was modified to support 200~Hz continuous position control, whereas the off-the-shelf version does not support real-time use. 

\subsection{VisTA Multimodal Policy}
\label{subsec:policy}
\begin{figure}[t]
    \centering
    \includegraphics[width=1.0\columnwidth]{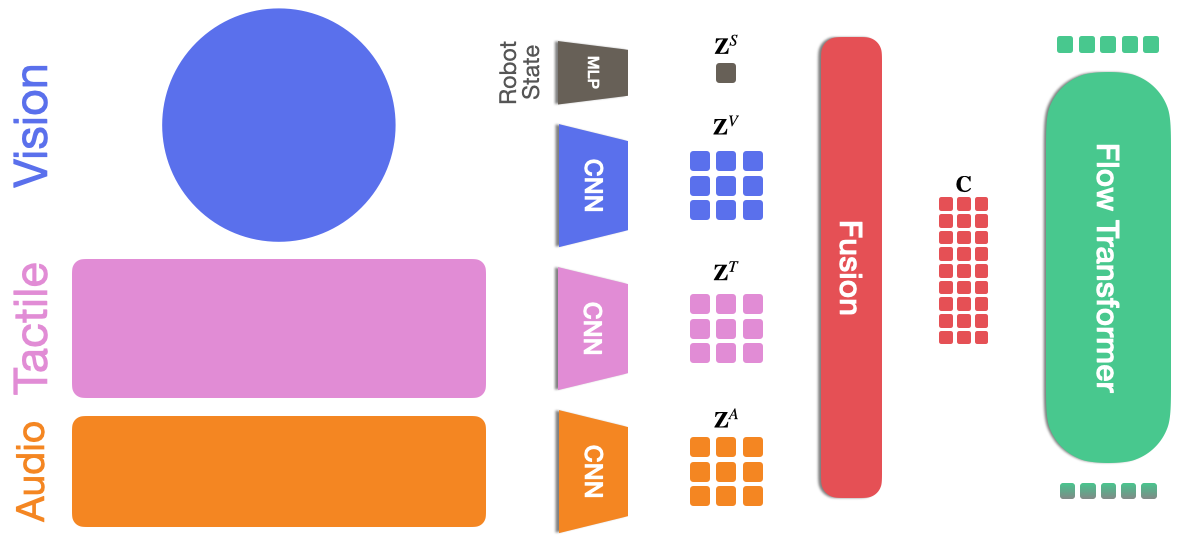}
    \caption{The VisTA model consists of sensor-specific CNN encoders and joint self-attention fusion. The high number of fused conditioning tokens $\mathbf{C}$, then conditions a flow matching transformer.}
    \label{fig:vista_model}
    \vspace{-0.5em}
\end{figure}

Observations obtained by PolyUMI provide complementary information at different spatial and temporal scales. Wrist vision captures the global scene, tactile images describe local contact geometry, contact audio captures short interaction events, and proprioception represents the robot state. Effectively combining these signals requires preserving their modality-specific structure while allowing information to be exchanged across sensors and time. We therefore propose \textbf{VisTA}, a token-level multimodal fusion architecture which enables the policy to dynamically combine global \textbf{Vis}ual context with local \textbf{T}actile and \textbf{A}uditory feedback in order to solve contact-rich manipulation tasks.
Given a small history of multisensory obervations 
\mbox{$\mathcal{O}_t =
    \left\{
        \mathbf{o}^{V}_{t-H+1:t},
        \mathbf{o}^{T}_{t-H+1:t},
        \mathbf{o}^{A}_{t},
        \mathbf{s}_{t-H+1:t}
    \right\},$} and the robot state $\mathbf{s}_t =
    \left[
        \mathbf{p}_t,\,
        \mathbf{r}_t,\,
        g_t,\,
        \mathbf{r}^{\mathrm{rel}}_t
    \right]
    \in \mathbb{R}^{16}$, the policy predicts an action chunk of $\mathbf{x}\in\mathbb{R}^{T_a\times d_a}$ with $T_a=16$ and $d_a=10$.
Each action contains a 3D translation, a 6D rotation, and a
relative gripper-width command. The pose commands are expressed relative to the current end-effector pose, making the action representation independent of the global
workspace frame. An Overview of the VisTA encoding and policy head is given in Figure~\ref{fig:vista_model}.

Each modality of the observation $O_t$ is processed to form embeddings of $D=624$. Wrist camera and tactile images are encoded by a CNN with a stride of 32, 
resulting in 98 tokens over $H=2$ timesteps. Modality-specific spatial embeddings and shared temporal embeddings retain the origin and time index of each token. 
The log-Mel spectrogram is encoded by an audio CNN to 96 tokens, while the robot state is independently encoded via an MLP.
The full observation 
$ \mathbf{Z}_0 =
    \left[
        \mathbf{Z}^{V};
        \mathbf{Z}^{T};
        \mathbf{Z}^{A};
        \mathbf{Z}^{S}
    \right]
    \in \mathbb{R}^{N\times D}$ leads to $N=294$ tokens. 
An eight-layer pre-normalized transformer with eight attention heads ($f_{\mathrm{fusion}}$) jointly refines this sequence by bi-directional self-attention and MLP subblocks. The attention mechanism operates over the complete token sequence, leading to direct information exchange across modalities, spatial locations, and observation times. In contrast to fusing one pooled feature per sensor, this design preserves local tactile deformation, short auditory events, and visual scene to condition the policy head.
The fused representation $\mathbf{C}= f_{\mathrm{fusion}}(\mathbf{Z}_0)$ conditions a DiT-style~\cite{peebles2023scalablediffusionmodelstransformers} action model with 18
transformer layers via cross attention. The model predicts a conditional velocity field
$\hat{\mathbf{v}}_{\theta}\left( \mathbf{x}_{\tau}, \tau, \mathbf{C}\right)$
over the complete action chunk and is trained using conditional flow matching~\cite{lipman2023flowmatching, black2026pi0}. The continuous flow time $\tau$ is injected into each DiT block using AdaLN-Zero conditioning~\mbox{\cite{dasari2024ingredientsroboticdiffusiontransformers}}.

\section{Experiments}
We evaluate PolyUMI along three claims that motivate our system. 
First, we ask whether its tactile and contact-audio streams can be used to expose object properties that are difficult to infer from wrist vision alone.
Second, we test whether these modalities provide precise feedback for closed-loop slip control. 
Third, we evaluate whether the resulting multimodal observations improve imitation policies on contact-rich manipulation tasks.
We use V, T, and A to denote wrist vision, optical tactile images, and contact audio, respectively.

\subsection{Experimental Procedure}
\label{subsec:experimental_procedure}

For all experiments, we use a Franka FR3 equipped with our PolyUMI end-effector. Demonstrations and policy rollouts therefore share the same sensing geometry.
Policies predict an action chunk containing 16 future actions. The first three actions are executed with temporal ensemble~\cite{zhao2023learningfinegrainedbimanualmanipulation} before the policy receives a new observation and replans. Within each experiment, all baselines and modality ablations use 
the same demonstrations, train--test split, observation and action horizons, input preprocessing, data augmentation, and optimization budget.
We train every model for 120 epochs and select the final checkpoint for evaluation. For manipulation and control tasks, we report the number of successful rollouts over the total number of evaluation trials. For classification tasks, we report accuracy on the held-out test set.

\subsection{Object Properties and Slip Control}
\label{subsec:object_properties}
We first evaluate whether PolyUMIs modalities capture task-relevant information arising during contact, before assessing their impact on longer-horizon manipulation.


\begin{figure}[t!]
    \centering
    \includegraphics[width=1.0\columnwidth]{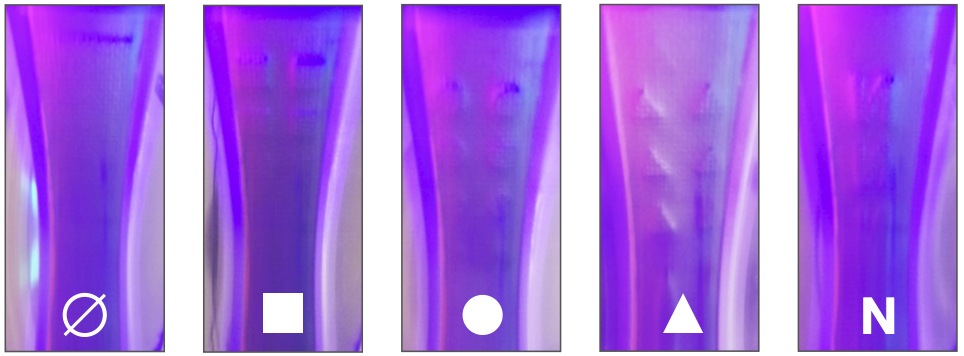}
    \caption{Tactile images of the shape recognition objects. From left to right: \textit{Flat}, \textit{Square}, \textit{Circle}, \textit{Triangle}, and the letter \textit{N}.}
    \label{fig:texture_objects}
\end{figure}

\subsubsection{\textbf{Tactile Shape Recognition}}
\begin{figure}[t]
    \centering
    \includegraphics[width=1.0\columnwidth]{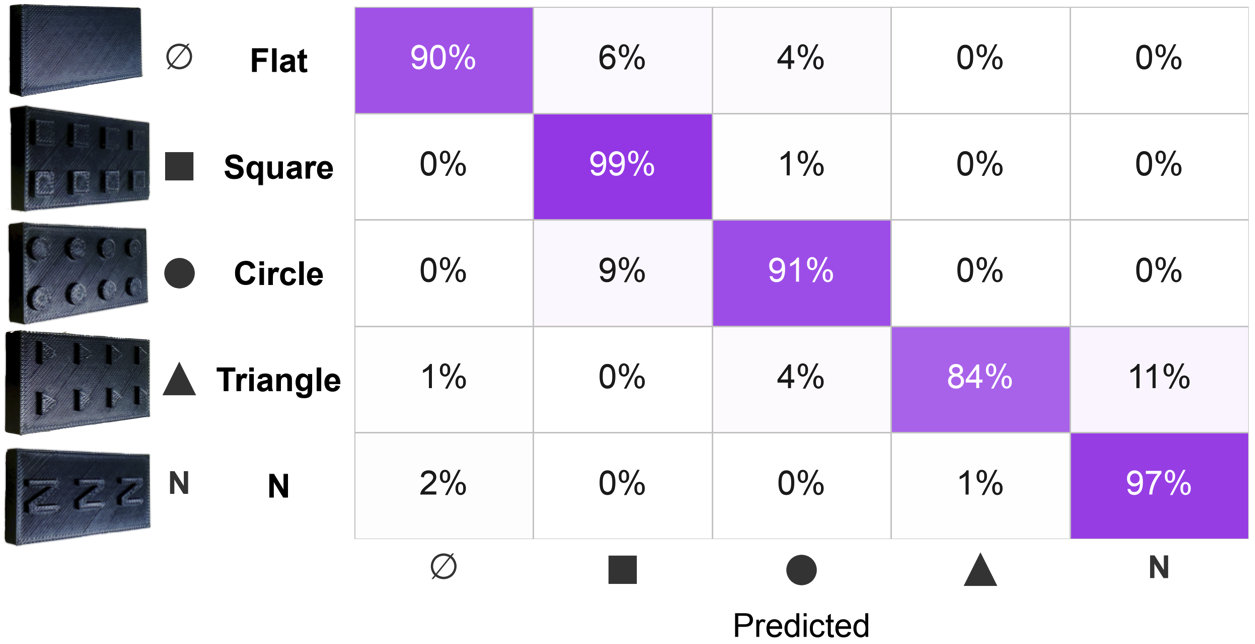}
        \vspace{-1.5em}
    \caption{Tactile shape classification results. PolyUMIs tactile sensor enables object recognition across 5 classes.}
    \label{fig:texture_results}
    \vspace{-1.0em}
\end{figure}

Fine surface geometry is difficult to recover visually during manipulation, as the contact
region is occluded by the gripper and the relevant features are smaller than the effective
resolution of an external camera.
We therefore investigate whether PolyUMI's tactile sensor captures geometry-specific
deformation signatures that are sufficiently informative to distinguish surface features at the contact interface.
We formulate this as a classification task over five 3D-printed patterns, i.e., \textit{Flat}, \textit{Square}, \textit{Circle}, \textit{Triangle}, and the letter \textit{N}, and train a classifier to predict the pattern from a single tactile reading.
The tactile readings are shown in Fig.~\ref{fig:texture_objects} while the results are displayed in Fig.~\ref{fig:texture_results}.


We collect 27 scenes per class under varying backgrounds, lighting conditions, and in-hand
object poses. Each scene contributes approximately 20 samples, resulting in a total of 135 unique settings and 2700 tactile images. We split the data at the scene level: the classifier is trained on 1986 samples from 20 scenes per class and evaluated on the 714 samples from the 7 held-out scenes per class, ensuring that no scene appears in both splits. The model consists of a CNN-based tactile encoder followed by an MLP head that predicts the class label from a single tactile image. Training runs for 120 epochs and takes approximately 11 minutes on a GeForce RTX 4060. Fig.~\ref{fig:texture_results} reports the resulting confusion matrix. The classifier reaches an accuracy of 92.3\,\%, indicating that PolyUMI's optical tactile sensor provides features that are sufficiently discriminative for tactile shape recognition.

\subsubsection{\textbf{Object-in-Box Classification}}
\label{subsubsec:object_in_box}

Many object properties remain initially hidden when objects need to be manipulated that are not fully hollow or filled, but contain objects inside. 
Humans can handle these scenarios and infer a container's contents by shaking it and listening to the resulting impacts. 
These sounds depend on properties such as the objects' mass, geometry, and material, whereas vision can observe only the motion of the container and robot. We therefore investigate whether PolyUMI's contact-audio sensor captures object-specific collision signatures that are sufficiently informative to classify visually occluded objects.

\begin{figure}[t]
    \centering
    \includegraphics[width=1.0\columnwidth]{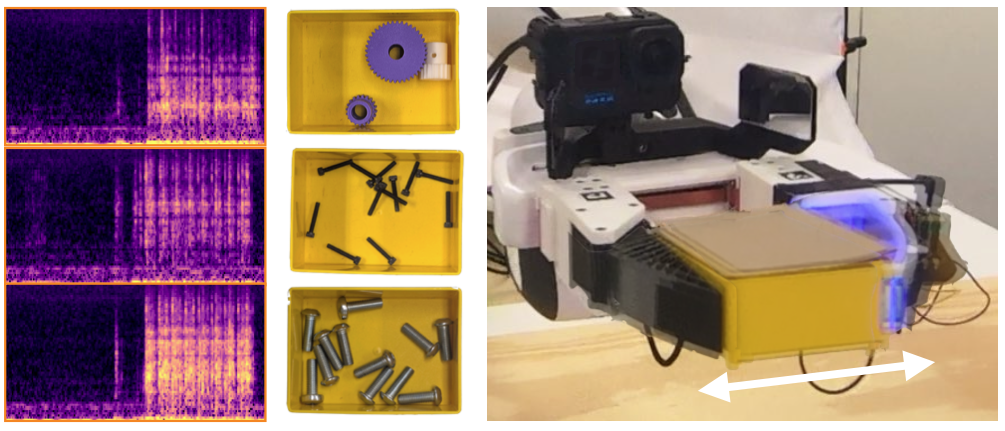}
    \caption{Audio signals (left) generated by shaking gears, thin screws, and thick screws (middle) with the robot arm (right) enable classification of visually occluded objects, substantially outperforming vision and touch.}
    \label{fig:task_setup_shake}
    \vspace{-0.5em}
\end{figure}

Inspired by~\cite{xu2025multimodaltactilefingertipdesign}, we place one of three object classes, i.e., gears, thin screws, or thick screws, inside the same closed box.
At the beginning of each demonstration, the box position is randomized over a 6\,cm range along the $x$-, $y$-, and $z$-axes. As shown in Fig.~\ref{fig:task_setup_shake}, the robot then shakes the box for 20 oscillations by following a sinusoidal trajectory with an amplitude of 10\,cm and a period of 0.15\,s. This controlled motion allows us to compare the sensory signals generated by the different object classes while varying the initial configuration of the objects inside the box.


We collect 30 demonstrations for each object class, using 20 demonstrations for training and reserving 10 for validation. Each available modality is processed by a separate CNN encoder. The resulting modality features are concatenated and passed to an MLP classification head that predicts one of the three object classes. To account for variability during training, we repeat each experiment with three random seeds and select the checkpoint with the highest validation accuracy. The confusion matrices therefore aggregate 30 predictions per object class and 90 predictions in total for each sensor configuration.

\begin{figure}[h]
    \centering
    \vspace{-0.5em}
    \includegraphics[width=\columnwidth]{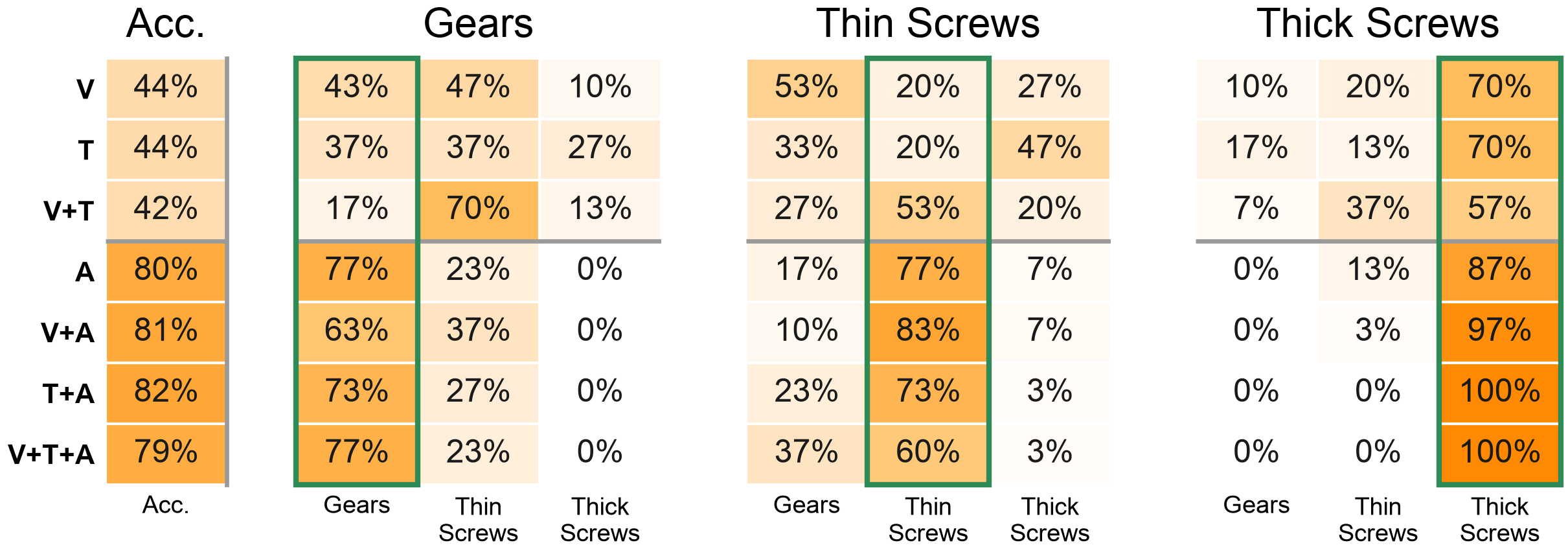}
    \caption{Prediction distributions for object-in-box classification,
    aggregated over three random seeds. Each panel corresponds to one
    ground-truth object class, and each row represents a sensor configuration.
    The green outlined cells indicate correct predictions.}
    \label{fig:shake_results_all}
    \vspace{-0.5em}
\end{figure}

Figure~\ref{fig:shake_results_all} shows that models without audio, i.e., vision-only ($44\,\%$), tactile-only ($44\,\%$) and visual--tactile ($42\,\%$), achieve only $10\,\%$ higher accuracy than random guessing ($33\,\%$). 
Notably, the vision-only model identifies thick screws more reliably than the other classes, reaching $70\,\%$ accuracy. This suggests that the model can partially exploit differences in the visible arm and container motion caused by the greater mass of the thick screws. However, visual observations alone do not provide sufficient information to distinguish the three classes reliably. Consequently, adding contact audio substantially improves classification leading to~$\approx80.0\,\%$ in every sensor configuration. 
In particular, the two tactile-audio conditioned models classify thick screws with $100\,\%$ accuracy, while The visual--audio model correctly identifies thin screws in $83\,\%$ of the trials. These results demonstrate that the structured audio recorded by PolyUMI contains object-specific information that is not available from vision or tactile images alone and  enables the classification of objects that remain occluded inside the box with only 20 training demonstrations per class.



\subsubsection{\textbf{Closed-Loop Slip Control}}
Finally, successful manipulation requires the robot to also respond to contact events as they evolve. In this context, object slippage is particularly challenging to control because it can be subtle, visually occluded, or too rapid to detect reliably from wrist-camera observations alone. Tactile deformation and structure-borne vibrations, in contrast, provide direct measurements of relative motion between the object and gripper. 
We therefore investigate whether PolyUMI’s tactile and contact-audio modalities provide sufficiently precise feedback for closed-loop slip control.

\begin{figure}[t]
  \centering
  
    \noindent
    \begin{minipage}[t]{0.30\columnwidth}
      \vspace{0pt}
      \centering
      \renewcommand{\arraystretch}{1.1} 
      \begin{tabular}[t]{cc}
          \hline
      \multicolumn{2}{c}{Slip Control} \\
        \hline
        Sensors & SR  \\
        \hline
        V+T+A  & \textbf{8/10} \\
        T+A  & 7/10  \\
        V+T  & 3/10  \\
        V+A  & 3/10 \\
        A  & 0/10  \\
        T  & 3/10  \\
        V  & 2/10  \\
        \hline
      \end{tabular}
      \vspace{-3.0pt}
        \label{tab:slip_control_results}
    \end{minipage}%
    \hspace{0.06\columnwidth}%
    \begin{minipage}[t]{0.64\columnwidth}
      \vspace{0pt}
      \centering
        \includegraphics[width=1.0\columnwidth]{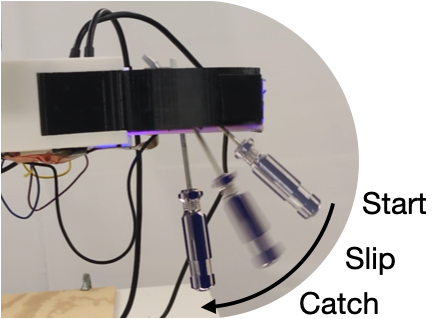}

      \vspace{0.0em}

    \end{minipage}
            \caption{Sensor ablation success rates (SR) across 10 trials (left) and side view of the Slip Control Task setup (right). The combination of tactile and audio sensing enables closed-loop slippage control, where other sensor modalities fail.}
              \label{fig:slip_control}
                  \vspace{-0.5em}
\end{figure}



In this experiment, the robot initially holds a screwdriver approximately horizontal, at $90^{\circ}$ from the downward direction~(see Fig.~\ref{fig:slip_control}).
The policy's goal is to rotate the screwdriver to $0^{\circ}$ (approximately vertical) by modulating the grasping force, without allowing the screwdriver to fall.
This task requires a careful balance, as tightening too early prevents the screwdriver from reaching the target orientation, whereas reacting too late causes the object to slip out of the gripper. Performing this task via teleoperation without haptic feedback is particularly hard due to the fine motor control required, making our handheld PolyUMI platform essential to train a successful policy.
In this experiment, a rollout is considered successful if the screwdriver remains in the gripper with a final orientation between $0^{\circ}$ and $10^{\circ}$ from the goal position.

For this task, we collect 80 demonstrations and encode the visual and tactile frames using ResNet backbones, while contact audio is processed using a temporal CNN.
The resulting features condition a diffusion-policy U-Net that predicts relative changes in gripper width.
By comparing the full visual--tactile--audio policy with different modality ablations, we assess whether contact sensing enables more accurate and timely regulation of object slip.
The full V+T+A policy succeeds in 8 out of 10 trials, compared to 2 out of 10 trials for the vision-only policy, as shown in Fig.~\ref{fig:slip_control}. 
These results suggest that contact-based observations provide important feedback for controlling fine-grained gripper--object interactions.

\subsection{Contact-Rich Manipulation}
\label{crm}
The preceding experiments isolate each modality of our PolyUMI platform and highlight that it can perceive object properties and detect dynamic contact events. However, the central question is whether these multimodal observations also improve complete manipulation behaviors. In contact-rich tasks, a policy must transition between free-space motion and physical interaction, maintain appropriate contact, and respond to small changes that may be difficult to observe visually. Errors during these phases can accumulate over time, causing otherwise successful trajectories to fail during alignment, insertion, or surface following. We therefore evaluate whether the tactile and auditory information captured by PolyUMI improves end-to-end manipulation reliability.

    

\begin{figure}[t]
    \centering
    \setlength{\tabcolsep}{5pt}
    
    \begin{tabular}{cc}
    \includegraphics[width=0.52\columnwidth, trim= 0 20 20 0, clip]{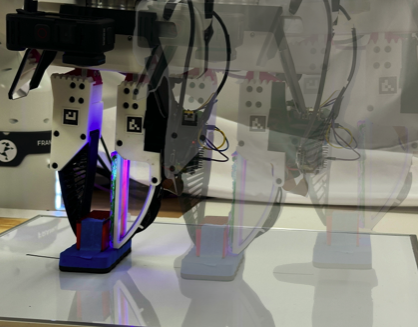} &
    \includegraphics[width=0.40\columnwidth, trim= 0 38 0 85, clip]{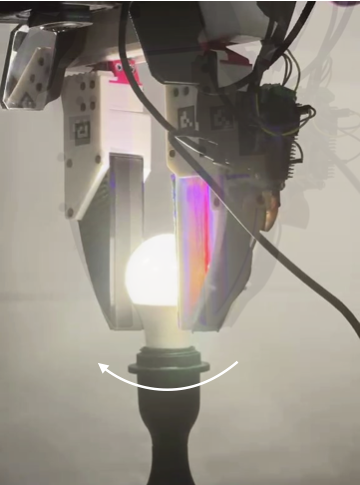}
    \end{tabular}
    \caption{Our contact-rich manipulation tasks Board Wiping (left) and Lightbulb Turning (right).}
    \label{fig:task_images}
    \vspace{-0.5em}
\end{figure}

Here, we consider two tasks that represent distinct forms of physical interaction. First, a \textbf{board wiping} task requires sustained contact and continuous surface following. Second, a lightbulb connection task combines the three difficult behaviors of alignment, insertion, and rotational engagement. All policies receive proprioceptive information together with the sensor observations. An overview of the task setups can be seen in Figure~\ref{fig:task_images}.
Together, these tasks allow us to evaluate whether multimodal policies can use contact information across different interaction types rather than only in a single controlled setting.

\subsubsection{\textbf{Baselines and Implementation}}
To determine whether performance gains arise from the additional contact modalities or from the way in which they are fused, we compare VisTA against both a vision-only policy and three multimodal fusion baselines.
The vision-only \textbf{Diffusion Policy}~\cite{chi2024universal,chi2025diffusion} uses a ResNet visual encoder and a U-Net policy head, providing a reference for manipulation without tactile or auditory observations. The remaining baselines receive the same visual, tactile, auditory, and proprioceptive inputs as VisTA but employ different strategies for encoding and combining them.
\textbf{MulSA}~\cite{li_see_2022} processes each modality with a separate~\mbox{ResNet-18} encoder~\cite{he2015deepresiduallearningimage} and applies attention across modalities and time. We replace the original MLP prediction head with a U-Net diffusion model to match other baselines.
\textbf{Sparsh-X}~\cite{higuera_tactile_2025} represents the sensor observations as tokens using modality-specific patch stems and exchanges information through attention bottlenecks~\cite{nagrani2022attentionbottlenecksmultimodalfusion}. The fused tokens are summarized through attention pooling~\cite{chen2023contextautoencoderselfsupervisedrepresentation} and passed to a DiT policy head trained using flow matching~\cite{black2026pi0}.
\textbf{PolyTouch}~\cite{zhao2025polytouch} instead relies on pretrained modality-specific representations, using CLIP for vision~\cite{radford2021learningtransferablevisualmodels}, T3 for tactile sensing~\cite{zhao_transferable_2024}, and an Audio Spectrogram Transformer for contact audio~\cite{gong_ast_2021}. Its features are combined through cross-attention and concatenated with the audio class token before conditioning the diffusion policy head.
This controlled comparison allows us to separately assess the value of contact sensing and the effect of the multimodal fusion architecture.

\subsubsection{\textbf{Board Wiping}}
Board wiping tests whether a policy can establish and maintain sustained contact while following a visually defined trajectory. Although the line to be erased is visible, successful execution cannot be achieved through visual tracking alone, as the policy must also regulate the interaction between the eraser and the board throughout the motion. 
Insufficient contact leaves parts of the line unerased, whereas deviations from the line produce only partial task completion. 
This task therefore evaluates whether multimodal contact observations help the policy coordinate contact maintenance with visually guided surface following.
In each trial, the robot must erase a line of~$\approx 25$~cm in length. 
The position of the line is slightly randomized between trials, while the robot begins from the same initial configuration.
We consider a rollout fully successful if the complete line is erased and partially successful if the robot maintains contact with the board but fails to erase the entire line. Losing board contact during the task is considered a failure.

\begin{figure}[t]
    \centering
    \includegraphics[width=1.0\columnwidth]{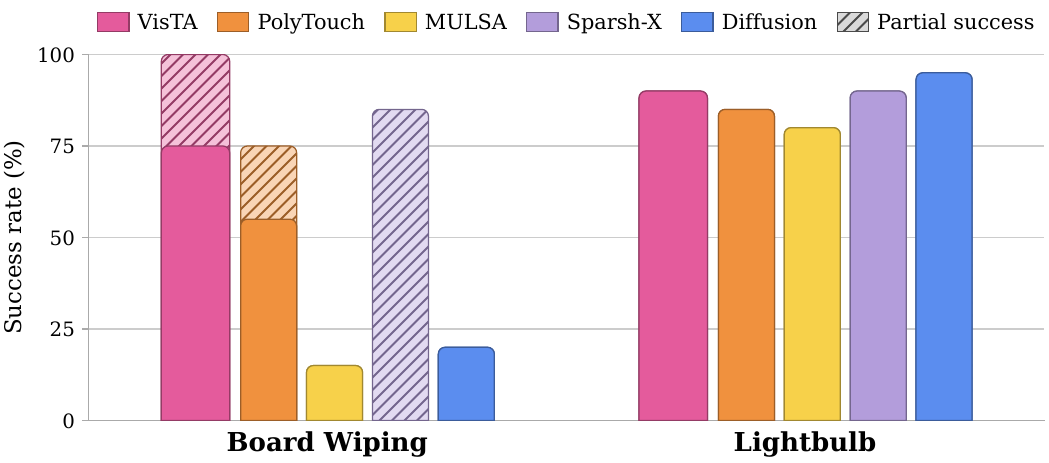}
    \caption{Manipulation results of Board Wiping and Lightbulb. VisTA outperforms state of the art visual tactile audio models in the board wiping task. All models are able to solve Lightbulb turning. while multisensor models fail to detect task stages, rather than contact modes.}
    \label{fig:manipulation_results}
    \vspace{-0.5em}
\end{figure}

The results are shown in Fig.~\ref{fig:manipulation_results}. The vision-only Diffusion Policy and MulSA frequently lose contact with the board during the wiping motion, leaving substantial portions of the line unerased. Sparsh-X generally maintains contact but often deviates from the line, resulting primarily in partial successes. In contrast, VisTA reliably establishes contact, maintains it throughout the wiping motion, and follows the line closely enough to erase it completely. These results indicate that successful board wiping requires both contact regulation and spatial tracking, and that VisTA effectively integrates the corresponding tactile, auditory, and visual information.

\subsubsection{\textbf{Lightbulb Turning}}
Lightbulb turning evaluates whether a policy can coordinate grasping, axial contact, and rotational motion across a multi-stage manipulation sequence. In each trial, the robot begins with a loosely inserted lightbulb and must rotate it until it is fully seated and switches on. We randomize the required bulb rotation and vary the initial vertical gripper position by $\pm 1$\,cm. The policy must maintain a stable grasp, reposition the gripper when necessary, and regulate the bulb's axial position while rotating it. A rollout is considered successful if the bulb switches on within 90\,s and remains seated after the gripper releases it.

As shown in Fig.~\ref{fig:manipulation_results}, all methods achieve success rates of at least $80\%$. The vision-only Diffusion Policy performs best, closely followed by VisTA and Sparsh-X. The remaining failures occur primarily during transitions between rotation stages. MulSA occasionally loses the bulb while re-grasping, whereas PolyTouch sometimes retreats before the bulb is fully seated. These results indicate that the task is largely solvable from visual observations because the bulb state and task completion provide clear visual cues. Consequently, tactile and auditory sensing offer limited additional benefit in this setting. Nevertheless, VisTA matches the strongest multimodal baseline and remains close to the vision-only policy, showing that its multimodal fusion supports multi-stage manipulation without compromising performance when vision provides the dominant task information.
\section{Conclusion}
In this work, we introduced PolyUMI, an open-source, wireless platform for synchronized visual--tactile--audio demonstration collection and robot deployment. By sharing the same sensing finger between handheld and robot-mounted embodiments, PolyUMI preserves the sensing configuration while enabling data collection without a tethered workstation. We further proposed VisTA, a token-level multimodal policy that combines information across sensors and time to generate contact-aware actions. Experiments on object-property inference demonstrated that tactile observations can be used to capture surface geometry, while contact audio reveals hidden object properties. Sensor ablations further showed that combining tactile and auditory feedback improves closed-loop slip control over vision alone. Finally, real-robot manipulation experiments demonstrated that VisTA outperforms the evaluated baselines on board wiping by coordinating sustained contact with visual tracking, while matching the strongest multimodal baseline on lightbulb turning. Together, these results highlight the complementary value of contact sensing and effective multimodal fusion for learning manipulation skills beyond purely visual feedback.



\section{Acknowledgments}
\label{sec:acknowledgements}
We thank Zhengxiao Han for his invaluable advice at the outset of this project, and Toby Buckley and Yutao He for insightful conversations throughout. This research is funded by the German Research Foundation (DFG) Emmy Noether Programme (CH2676/1-1), the EU’s Horizon Europe project ARISE (Grant no.: 101135959), the German Federal Ministry of Education and Research (BMBF) project “RiG” (Grant no.: 16ME1001), the European Research Council (ERC) project “SIREN” (Grant No.: 101163933), the National Science Foundation (Grant No.: 2221571 and Grant No.: DGE-2234667), and the Northwestern University MS in Robotics program. The authors gratefully acknowledge the scientific support and HPC resources provided by the Erlangen National High Performance Computing Center (NHR@FAU) of the Friedrich-Alexander-Universität Erlangen-Nürnberg
(FAU), and Northwestern University's Quest High Performance Computing Cluster.
\newpage

\bibliographystyle{IEEEtran}
\bibliography{references}

\end{document}